\documentclass[runningheads]{llncs}

\usepackage[T1]{fontenc}

\usepackage{graphicx}

\usepackage{amsmath}
\usepackage{amsfonts}
\usepackage{multirow}
\usepackage{booktabs}
\usepackage{xcolor}
\usepackage{enumerate}

\newcommand{\Struct}{\mathbf{S}}
\newcommand{\Neigh}{\mathbf{N}}

\begin{document}
\title{Drop the mask! GAMM - A Taxonomy for Graph Attributes Missing Mechanisms}
\titlerunning{Drop the mask! GAMM}
%
\author{Richard Serrano\inst{1} \and\orcidID{0009-0009-4946-896X} \and\\
    Baptiste Jeudy\inst{1}\orcidID{0009-0000-8126-2608} \and\\
    Charlotte Laclau\inst{2}\orcidID{0000-0002-7389-3191} \and\\
    Christine Largeron\inst{1}\orcidID{0000-0003-1059-4095}}
\authorrunning{R. Serrano et al.}
%
\institute{Laboratoire Hubert Curien, Saint-Étienne 42000, France\\
    \email{\{richard.serrano; baptiste.jeudy; christine.largeron\}@univ-st-etienne.fr} \and
    Télécom Paris, Institut Polytechnique de Paris, France\\
    \email{charlotte.laclau@telecom-paris.fr}}

\maketitle              
\begin{abstract}
\begin{sloppypar}
    Exploring missing data in attributed graphs introduces unique challenges beyond those found in tabular datasets. In this work, we  extend  the taxonomy for missing data mechanisms to attributed graphs by proposing GAMM (Graph Attributes Missing Mechanisms), a framework that systematically links missingness probability to both node attributes and the underlying graph structure. Our taxonomy enriches the conventional definitions of masking mechanisms by introducing graph-specific dependencies.\\
    We empirically demonstrate that state-of-the-art imputation methods, while effective on traditional masks, significantly struggle when confronted with these more realistic graph-aware missingness scenarios.
\end{sloppypar}

\end{abstract}

\section{Introduction}\label{sec:intro}
Attributed graphs provide a powerful framework for modeling complex systems, from social networks to biological interactions. Unlike tabular data, graph-structured data captures the inherent connections between entities, where a node is defined by both its attributes and its relational context. However, this structure creates unique challenges when dealing with missing data. In practice, node attributes are often partially unobserved, posing a significant problem for methods like Graph Neural Networks (GNNs) that usually require fully observed feature matrices. For instance, in social networks, users may omit sensitive profile information depending not only on their own characteristics but also on their position within the network or the attributes of their friends. A user with many connections might be more cautious about sharing personal data, or missingness may cluster within communities due to shared privacy norms or platform features. Such dependencies between missingness and network structure are not captured by classical missing data models, which assume independence from relational context. Accurately modeling these effects is crucial for fair and effective inference on graph data. This need has led to substantial research into missing attribute imputation \cite{rossiUnreasonableEffectivenessFeature2022,um2023confidence,serrano2024reconstructing}. However, a critical issue is that these methods are often evaluated using simplistic masking mechanisms, e.g., MCAR, MAR, MNAR \cite{liLittlesTestMissinga}, which provides a crucial foundation, but do not explicitly account for structural dependencies. In real-world graphs, missingness may be influenced by a node's connectivity or the characteristics of its neighborhood.

\paragraph{Contributions.}
\begin{sloppypar}
    We introduce the Graph Attributes Missing Mechanisms (GAMM) taxonomy, which extends traditional missingness frameworks by incorporating structural dependencies and classifying missingness into four sources: (A) node attributes, (S) structural properties, (N) neighbors’ attributes, and (G) their combinations. Focusing solely on missing node attributes under a known topology, we show that graph-aware mechanisms substantially degrade the performance of state-of-the-art imputers relative to attribute-only missingness. To structure the investigation, we examine how (Research Question 1) S- and N-based mechanisms affect reconstruction performance across graphs and missing rates, (RQ2) homophily/heterophily modulates the gap between attribute- and neighborhood-based MAR, (RQ3) neighbor-dependent MNAR compares to attribute-based MNAR and in which settings it becomes most severe, and (RQ4) whether imputers preserve feature distributions under these structurally informed conditions. An open-source implementation of all mechanisms and protocols is provided.\footnote[1]{\label{ft:code}Code and Supplementary Material: \url{github.com/RichardSrn/GAMM}}
\end{sloppypar}

This paper is structured as follows. Section~\ref{sec:related_work} outlines existing masking strategies and discusses their limits. Section~\ref{sec:proposed_taxonomy} presents GAMM; Section~\ref{sec:experiments} details experiments and results. Finally,  Section~\ref{sec:conclusion} concludes.

\section{Limits of Traditional Missing Data Mechanisms}\label{sec:related_work}

\subsection{Traditional Missing Data Mechanisms}

The traditional taxonomy identifies three missingness mechanisms \cite{rebininference1976,littleTestMissingCompletely1988}: (1) \textbf{Missing} \textbf{Completely At Random (MCAR)}, where missingness is independent of all data; (2) \textbf{Missing At Random (MAR)}, where it is influenced only by observed data; and (3) \textbf{Missing Not At Random (MNAR)}, where it depends on unobserved values. For instance, in a corporate network, missing employee \textit{Salary} data would be MCAR if it resulted from a randomly occurring glitch in the employee database, MAR if junior employees (i.e., observed \textit{Experience} attribute) are less likely to report it, and MNAR if employees with higher salaries (the unobserved feature itself) are the ones who mostly withhold this information. Most imputation methods assume MCAR or MAR conditions \cite{little2024comparison}, while MNAR requires specialized approaches \cite{hsu2020multiple}.
Existing extensions to this taxonomy, including graphical models \cite{mohan2021graphical} and conditional independence approaches \cite{doretti2018missing}, enhance the framework for tabular data but fail to address network-specific challenges where structural relationships directly influence missingness patterns. Similarly, recent work advocating for explicit measurement process modeling \cite{lee2023assumptions} acknowledges the limitations of the rigid MCAR/MAR/MNAR classification in complex dependencies, yet still overlooks how network topology itself can determine whether attributes are observed. While these foundational missingness mechanisms provide a basis for our graph-based taxonomy, they require significant extensions to address the unique complexities of network structure.

\subsection{Taxonomies and Extensions for Graph-Based Missingness}

When translating missingness concepts to graphs, structural relationships between entities add complexity beyond tabular settings. Missingness can depend on node attributes, topological properties, or relational influences \cite{kossinetsEffectsMissingData2006,huismanImputationMissingNetwork2014}.

\paragraph{Limitations.}
We claim that traditional definitions of missingness mechanisms are inadequate for graph-structured data, where missingness may depend on structural properties. The following two examples illustrate this argument:

\begin{enumerate}
    \item \textbf{Structurally-Induced Missingness}: Nodes with specific structural properties (e.g., low degree) may exhibit missing attributes due to their position in the network. In social networks, peripheral users may be less likely to report certain attributes, like alcohol consumption \cite{huismanImputationMissingNetwork2014}. This represents a MAR mechanism dependent on graph topology rather than attributes values; a distinction classical frameworks fail to characterize.
    \item \textbf{Neighbor-Influenced Missingness}: In social networks, individuals might withhold information (e.g., salary) not based on their own values, but influenced by their friends' values. This creates dependencies unexplainable without incorporating graph structure \cite{kossinetsEffectsMissingData2006}. This scenario represents a graph-specific form of MNAR because the missingness of a node's attribute is conditioned on the unobserved attributes of its neighbors. Here, traditional MNAR frameworks are inadequate, as they are unable to model dependencies that arise from the underlying graph topology. This concept of leveraging network information can also be applied to the MAR framework, for instance, by conditioning the missingness on the \textit{observed} features of neighboring nodes.
\end{enumerate}

In the following, we propose a new taxonomy for attributed graphs, providing a basis for clear methodology and consistent evaluation of graph-imputation methods under more realistic missingness conditions.

\section{Taxonomy for Graph Attributes Missing Mechanisms (GAMM)}\label{sec:proposed_taxonomy}

In this section, we introduce GAMM (Graph Attributes Missing Mechanisms), a novel taxonomy for understanding and generating missing data masks in attributed graphs. GAMM extends traditional missingness definitions by explicitly incorporating the interplay between node attributes and the graph structure.

\begin{figure*}[t]
\centering
\hfill
\begin{minipage}[c]{0.80\linewidth}
\begin{minipage}{0.32\linewidth}
    \centering
    \includegraphics[width=\linewidth]{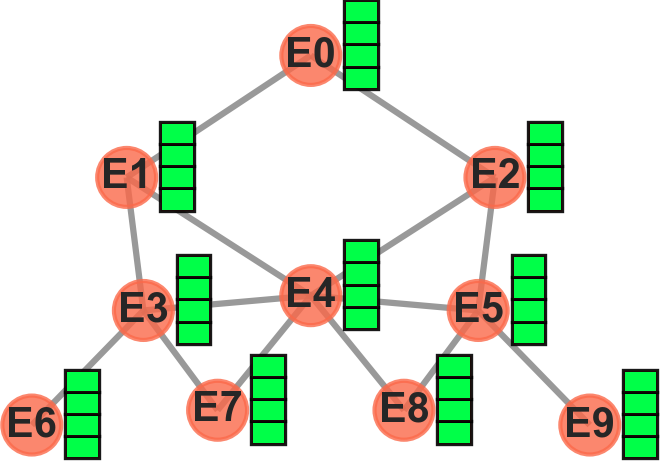}
    \centerline{\small(a) MCAR}
\end{minipage}%
\hfill
\begin{minipage}{0.32\linewidth}
    \centering
    \includegraphics[width=\linewidth]{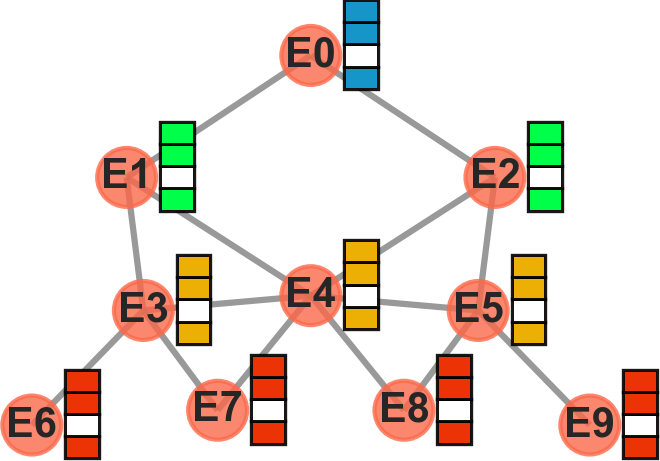}
    \centerline{\small(b) A-MAR}
\end{minipage}%
\hfill
\begin{minipage}{0.32\linewidth}
    \centering
    \includegraphics[width=\linewidth]{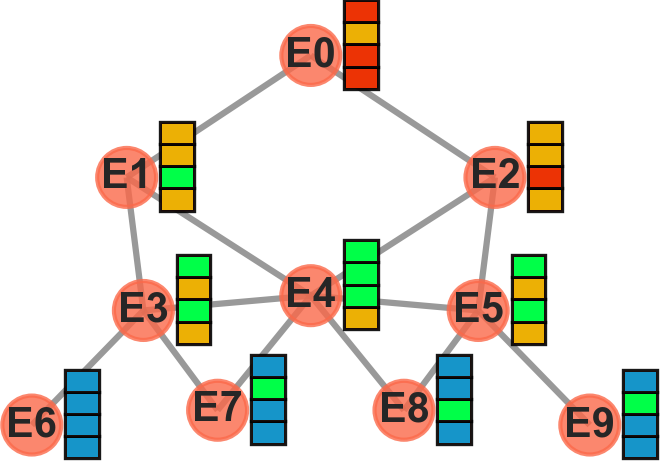}
    \centerline{\small(c) A-MNAR}
\end{minipage}

\vspace{0.5cm}

\begin{minipage}{0.32\linewidth}
    \centering
    \includegraphics[width=\linewidth]{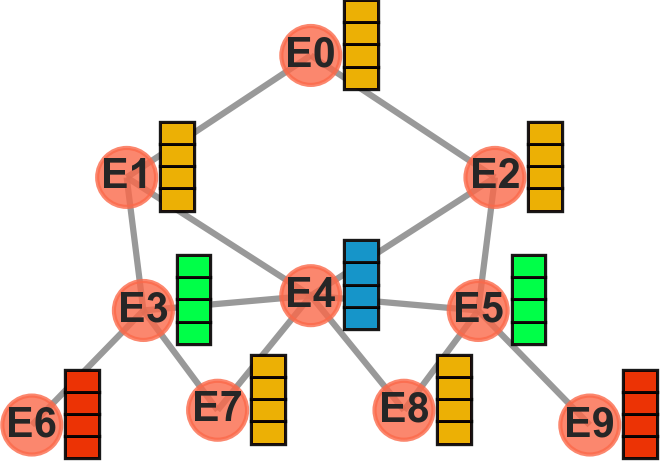}
    \centerline{\small(d) S-MAR}
\end{minipage}%
\hfill
\begin{minipage}{0.32\linewidth}
    \centering
    \includegraphics[width=\linewidth]{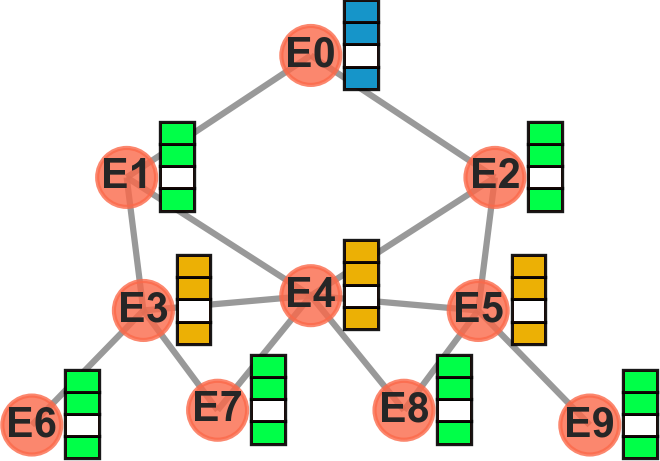}
    \centerline{\small(e) N-MAR}
\end{minipage}%
\hfill
\begin{minipage}{0.32\linewidth}
    \centering
    \includegraphics[width=\linewidth]{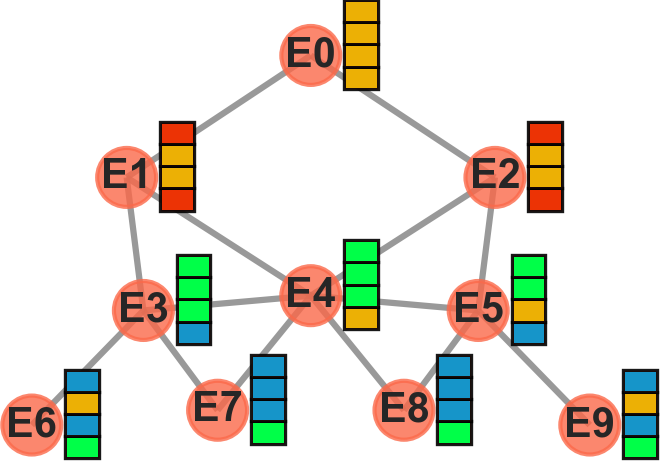}
    \centerline{\small(f) N-MNAR}
\end{minipage}
\end{minipage}%
\hfill
\begin{minipage}[c]{0.15\linewidth}
\vspace{0cm} 
\centering
\includegraphics[width=\linewidth]{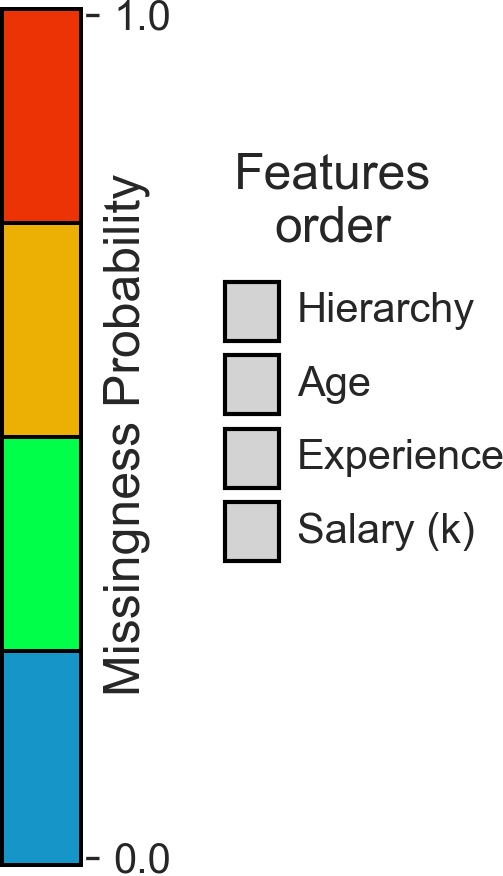}
\end{minipage}
\hfill

\caption{Traditional (a–c) and graph-aware (d–f) missingness mechanisms in GAMM. Each node shows four attribute slots (top-to-down: Hierarchy, Salary, Experience, Age); red shades indicate higher missingness probability, while blue indicates lower probability (white = observed). Consistent with the example description in paragraph \textit{Example}, higher positions in the hierarchy correspond to higher attribute values.}

\label{fig:missingness_mechanisms}
\end{figure*}

To illustrate these mechanisms, we use a toy example of a corporate email network (Figure~\ref{fig:missingness_mechanisms}). Each node represents an employee, and edges denote communication. Node attributes include \textit{Hierarchy}, \textit{Salary}, \textit{Experience}, and \textit{Age}, with some features correlated to aid visual intuition. 
To ensure consistent interpretation across panels, each node displays four attribute slots in the fixed top-to-down order \textit{[Hierarchy, Salary, Experience, Age]}. 
We encode the missingness probability $p_{\text{miss}}$ using a color scale: \emph{white} denotes observed entries ($p_{\text{miss}}=0$), followed by \emph{blue} for $p_{\text{miss}} \in (0, 0.25]$, \emph{green} for $p_{\text{miss}} \in (0.25, 0.50]$, \emph{orange} for $p_{\text{miss}} \in (0.50, 0.75]$, and \emph{red} for $p_{\text{miss}} \in (0.75, 1.0]$.\\

We consider an undirected attributed graph $G=(\mathcal{V},\mathcal{E},F)$ with $n$ nodes, edges $\mathcal{E}\subset\mathcal{V}\times\mathcal{V}$, and feature matrix $F\in\mathbb{R}^{n\times d}$ where $F_{i\cdot}$ is the feature vector of node $v_i$ and $\Struct_i$ are its structural properties (degree, \dots).
Missing entries are indicated by a binary mask $\Omega\in\{0,1\}^{n\times d}$, with $\Omega_{ij}=1$ when feature $j$ of $v_i$ is observed. We decompose $F$ into $F^{(\text{OBS})}$ (observed features) and $F^{(\text{MIS})}$ (features subject to missingness).

\paragraph{Example.}\label{par:example}
Consider a corporate email network. In a \textbf{MAR} scenario, suppose \textit{Experience} is fully observed while \textit{Hierarchy}, \textit{Age} and \textit{Salary} may be missing features. This partitions the features into $F^{(\text{OBS})}=[\textit{Experience}]$ and $F^{(\text{MIS})}=[\textit{Hierarchy},\textit{Age},\textit{Salary}]$.
In contrast, in a \textbf{MNAR} scenario, we may have no fully observed features ($F^{(\text{OBS})}=\emptyset$), meaning all attributes fall into $F^{(\text{MIS})}$.

Crucially, missingness in graphs often correlates with structure. For instance, junior employees, often located in peripheral regions of the graph, may be more likely to have missing features. That would be a \textbf{MAR} scenario based on structure, and imputers that ignore this structural pattern may systematically overestimate their roles, yielding distributional distortions that standard attribute-only evaluations fail to capture.

Note that in Figure~\ref{fig:missingness_mechanisms}, higher positions in the hierarchy correspond to higher attribute values for visual interpretation.

\paragraph{Graph Context.}
\begin{sloppypar}
    To model this context, we denote by $\mathcal{N}_h(v_i)$ the $h$-hop neighbors of $v_i$. We define the structural and feature-based neighborhoods as follows:
    \begin{itemize}
        \item \textbf{Structure:} $\Struct_i^{(\text{OBS})}(h)=\{\Struct_k \mid v_k\in\mathcal{N}_h(v_i)\text{ and } \Struct_k \text{ fully observed}\}$ is the set of observed structural properties in the $h$-hop neighborhood, while $\Struct_i^{(\text{MIS})}(h)=\{\Struct_k \mid v_k\in\mathcal{N}_h(v_i) \text{ and } \Struct_k \text{ may be missing}\}$ denotes those that may be missing.
        \item \textbf{Features:} Similarly, $\Neigh_i^{(\text{OBS})}(h)=\{F_{k\cdot}^{(\text{OBS})}\mid v_k\in\mathcal{N}_h(v_i)\}$ represents the observed features of neighbors, and $\Neigh_i^{(\text{MIS})}(h)=\{F_{k\cdot}^{(\text{MIS})}\mid v_k\in\mathcal{N}_h(v_i)\}$ represents the potentially missing features.
    \end{itemize}
\end{sloppypar}
For $h=0$, $\Struct_i^{(\text{OBS})}(0)$ and $\Struct_i^{(\text{MIS})}(0)$ correspond to the properties of $v_i$ itself, abbreviated as $\Struct_i^{(\text{OBS})}$ and $\Struct_i^{(\text{MIS})}$. A full notation list is provided in Section~A.3 of the Supplementary Material.

\subsection{Proposed Mechanisms}\label{sec:proposed_definitions}
We are now ready to present GAMM, a Graph-Aware Missingness Mechanism framework, under the assumption that the graph structure $(\mathcal{V},\mathcal{E})$ 
is fully observed. The probability that an attribute $F_{ij}$ is missing ($P(\Omega_{ij} = 0)$) is governed by a function $g(.)$ over relevant inputs:
\begin{equation}
P(\Omega_{ij}=0\mid G)=P(\Omega_{ij}=0\mid\cdot)=g(\cdot).
\end{equation}
The nature of the inputs to $g(.)$ determines the missingness mechanism. Table~\ref{tab:overview} summarizes our taxonomy, and illustrative examples are shown in Figure~\ref{fig:missingness_mechanisms}.
In addition to the MCAR setting, we distinguish four key families of mechanisms that capture different sources of dependency in graphs.

\begin{table*}[!t]
\centering
\caption{GAMM Taxonomy: Characterization of graph-aware missingness mechanisms based on the inputs to the probability function $g(\cdot)$. $\checkmark$ indicates an input is potentially used by $g(\cdot)$. A stands for Attribute-based, S for Structure, N for Neighborhood, and G for Generic.}
\label{tab:overview}
\resizebox{\linewidth}{!}{
\begin{tabular}{lcccccccc}
    \toprule
    \multirow{2}{*}{Cat.}        & \multirow{2}{*}{Mech.} &                           \multirow{2}{15em}{\centering Formal Definition of }                           &                                                                           \multicolumn{6}{c}{Possible dependencies for $g(\cdot)$}                                                                            \\
    \cmidrule(lr){4-9}           &                        &              $P(\Omega_{ij}=0\mid G)$                                                                          & $F_{i\cdot}^{\text{(OBS)}}$ & $F_{i\cdot}^{\text{(MIS)}}$ & $\Struct_i^{(\text{OBS})}(h)$ & $\Struct_i^{(\text{MIS})}(h)$ & $\Neigh_i^{\text{(OBS)}}(h)$ & $\Neigh_i^{\text{(MIS)}}(h)$ \\ \midrule
    &          MCAR          &                                   $P(\Omega_{ij}=0)$                                   &                             &                             &                                    &                                    &                                    &                                    \\ \midrule
    \multirow{2}{*}{\textbf{A-}} &          MAR           &                  \multirow{2}{*}{$P(\Omega_{ij}=0 \mid F_{i\cdot})$}                   &         \checkmark          &                             &                                    &                                    &                                    &                                    \\
    &          MNAR          &                                                                                        &         \checkmark          &         \checkmark          &                                    &                                    &                                    &                                    \\ \midrule
    \multirow{2}{*}{\textbf{S-}} &          MAR           &                         \multirow{2}{*}{$P(\Omega_{ij}=0 \mid \Struct_i)$}                         &                             &                             &             \checkmark             &                                    &                                    &                                    \\
    &          MNAR          &                                      &                             &                             &             \checkmark             &             \checkmark             &                                    &                                    \\ \midrule
    \multirow{2}{*}{\textbf{N-}} &          MAR           &               \multirow{2}{*}{$P(\Omega_{ij}=0 \mid \Neigh_i(h))$}               &                             &                             &                                    &                                    &             \checkmark             &                                    \\
    &          MNAR          &                                                                                        &                             &                             &                                    &                                    &             \checkmark             &             \checkmark             \\ \midrule
    \multirow{2}{*}{\textbf{G-}} &          MAR           & \multirow{2}{*}{$P(\Omega_{ij}=0 \mid F_{i\cdot}, \Struct_i, \Neigh_i(h))$} &         \checkmark          &                             &             \checkmark             &                                    &             \checkmark             &                                    \\
    &          MNAR          &                                                                                        &         \checkmark          &         \checkmark          &             \checkmark             &             \checkmark             &             \checkmark             &             \checkmark             \\ \bottomrule
    &                        &                                                                                        &                             &                             &
\end{tabular}
}
\end{table*}

\paragraph{Missing Completely At Random (\textbf{MCAR})}
Missingness occurs independently of both node attributes and the graph structure: the missingness probability is the same constant for all entries of $F^{(\text{MIS})}$, as shown in Figure \ref{fig:missingness_mechanisms}(a).

\paragraph{Attribute-based Missingness (\textbf{A-})}\label{par:attributes_missingness}
In this setting, missingness for a node $v_i$ depends only on its own attributes, mirroring standard tabular assumptions. \textbf{A-MAR} arises when the probability of missingness depends on observed attributes of $v_i$ (e.g., in Figure~\ref{fig:missingness_mechanisms}(b), low observed \textit{Experience} correlates with more missing values). Conversely, under \textbf{A-MNAR}, missingness is driven by the node’s unobserved attributes (Figure~\ref{fig:missingness_mechanisms}(c) illustrates higher true \textit{Salary} leading to missing entries).

\paragraph{Structural Properties-aware Missingness (\textbf{S-})}\label{par:structure_missingness}
Here, missingness depends on the structural characteristics of $v_i$ (degree, centrality, etc.). Under \textbf{S-MAR}, it is governed by observed structural features (Figure~\ref{fig:missingness_mechanisms}(d) shows low-degree nodes having more missing data). Under \textbf{S-MNAR}, missingness depends on missing structural characteristics; for instance, nodes with missing edges in the graph may exhibit more missing values on features too.

\paragraph{Neighborhood-aware Missingness (\textbf{N-})}\label{par:neighbors_missingness}
In this case, missingness of $v_i$ depends on attributes of its $h$-hop neighbors (where the neighborhood is defined by the graph topology and not from the feature-space). Under \textbf{N-MAR}, it depends only on neighbors’ observed attributes (e.g., Figure~\ref{fig:missingness_mechanisms}(e), neighbors’ observed \textit{Experience}). Under \textbf{N-MNAR}, it depends on neighbors’ unobserved attributes, as in Figure~\ref{fig:missingness_mechanisms}(f).

\paragraph{Generic Combination Missingness (\textbf{G-})}\label{par:generic_missingness}
This category captures the most general and realistic forms of missingness. It allows dependencies across a node's own attributes (both observed and unobserved), structural properties, and neighborhood information.
Such cases reflect complex real-world patterns where missingness cannot be attributed to a single source but instead arises from intertwined local and structural factors. These scenarios often blur the line between MAR and MNAR, making the masking mechanism particularly hard to disentangle without additional assumptions or annotations.

\paragraph{Remark On the Identifiability of Missingness Mechanisms}\label{par:identifiability}

A central challenge in modeling missing data is distinguishing MAR from MNAR mechanisms, a task that is fundamentally unidentifiable without additional assumptions or external supervision. This difficulty arises because the same observed missingness patterns can often be explained by either observed or unobserved variables.

Consider a feature with missing entries, $F_{\cdot j}$, that is correlated with a fully observed feature $F_{\cdot k}$. If missingness depends on $F_{\cdot k}$, it falls under MAR; but if it instead depends on the unobserved values of $F_{\cdot j}$, the mechanism is MNAR. In practice, these explanations are statistically indistinguishable from observational data alone leading to intrinsic ambiguity in determining the true source of missingness. 
In the tabular settings, identifiability is often assumed under the i.i.d. sampling of rows (i.e., independent individuals) \cite{sportisse:tel-03722429}, though this assumption is stronger than strictly necessary \cite{doretti2018missing}. For graphs, this assumption breaks down: nodes are not independent. However, for identifying the masking mechanism, we do not require node-level independence. Instead, it suffices to assume independence across features, i.e., columns of $F$ are uncorrelated.

While this assumption is rarely satisfied in real-world data (whether graph-based or tabular), it enables clear identifiability in controlled experiments. By constructing synthetic data under known and independent feature distributions, we can isolate and verify the missingness mechanism. This is crucial when benchmarking models or validating imputation strategies under known ground truth.

\paragraph{Summary.}
GAMM provides a formal framework to characterize and simulate diverse missingness mechanisms in graph data, capturing dependencies on node attributes, structural properties, and neighborhood information. This enables the creation of realistic missing data scenarios and systematic evaluation of their impact on graph learning methods. In the next section, we use GAMM to assess the robustness of state-of-the-art imputer models across these mechanisms. Our results show that performance and reliability vary significantly with the missing data type, highlighting the importance of accounting for these distinctions in graph algorithm design and evaluation.

\section{Experiments}
\label{sec:experiments}

\subsection{Experimental Protocol}

Our experimental design directly addresses \textbf{RQ1--RQ4} by systematically varying missingness mechanisms (traditional vs.\ graph-aware), graph homophily, and missing rates, and evaluating tabular and graph-based imputers using MAE/ RMSE and distributional analysis.

\paragraph{Datasets.}
We select twelve benchmark real-world graph datasets \cite{yang2016revisiting,rozemberczki2021multi,tang2009social,platonov2023critical,craven1998learning} 
(detailed in Section A.1 of the Supplementary Material) with diverse network properties, notably varying levels of homophily, which is crucial for evaluating our proposed mechanisms. 

Table \ref{tab:dataset_summary} summarizes their characteristics and groups them into Homophilic, Neutral, and Heterophilic categories based on their adjusted homophily ($H_{adj}$) \cite{pei2020geom}. 

\begin{table*}[t]
\centering
\caption{Presentation of the benchmark datasets, categorized by their adjusted homophily ($H_{adj}$).}
\label{tab:dataset_summary}
\begin{tabular}{llccclc}
    \toprule
    & Dataset      & \#Nodes & \#Edges & \#Classes & Features           & $ H_{adj} $ \\ \midrule
    \multirow{3}{*}{Homophilic}   & Cora         &  2708   &  5278   &     7     & $ \{0,1\}^{1433} $ &    0.768    \\
    & PubMed       &  19717  &  44324  &     3     & $ [0,2]^{500} $    &    0.693    \\
    & CiteSeer     &  3327   &  4552   &     6     & $ \{0,1\}^{3702} $ &    0.678    \\ \midrule
    \multirow{5}{*}{Neutral}      & Chameleon    &  2277   &  18050  &     5     & $\{0,1\}^{2325}$   &    0.042    \\
    & Squirrel     &  5201   & 108536  &     5     & $\{0,1\}^{2089}$   &    0.030    \\
    & Minesweeper  &  10000  &  39402  &     2     & $\{0,1\}^{7}$      &    0.009    \\
    & Actor        &  7600   &  15009  &     5     & $\{0,1\}^{932}$    &    0.007    \\
    & Roman-empire &  22662  &  32927  &    18     & $[0,1]^{300}$      &   -0.045    \\ \midrule
    \multirow{4}{*}{Heterophilic} & Tolokers     &  11758  & 519000  &     2     & $[0,1]^{10}$       &   -0.188    \\
    & Wisconsin    &   251   &   257   &     5     & $\{0,1\}^{1702}$   &   -0.189    \\
    & Cornell      &   183   &   149   &     5     & $\{0,1\}^{1702}$   &   -0.219    \\
    & Texas        &   183   &   162   &     5     & $\{0,1\}^{1702}$   &   -0.425    \\ \bottomrule
\end{tabular}
\end{table*}

\paragraph{Missingness Mask Generation.}
We generate masks at overall rates of $p_{\text{miss}} \in \{20\%, 50\%, 80\%\}$ using distinct mechanisms from our taxonomy. As introduced in Section \ref{sec:proposed_taxonomy}, we define the function $g(\cdot)$ that maps specific informational inputs to a missingness probability, $P(\Omega_{ij}=0) = g(\cdot)$. The specific ways in which $g(\cdot)$ is determined, which instantiate mechanisms based on a node's own attributes (A-), its structure (S-), or its neighbors' attributes (N-), are defined as follows:

\begin{equation}\label{eq:operational}
\begin{aligned}
&\text{MCAR: } g() = p_{\text{miss}} \\[0.2em]
&\text{A-MAR/MNAR: }\;
g(F_{i\cdot}) =
\begin{cases}
    \sigma(\omega_1 F_{i\cdot}^{(\text{OBS})}+b_1), & \text{A-MAR},  \\[-0.2em]
    \sigma(\omega_2 F_{ij}^{(\text{MIS})}+b_2),     & \text{A-MNAR},
\end{cases} \\[0.3em]
&\text{S-MAR/MNAR: }\;
g(\Struct_i)=
\begin{cases}
    \sigma(\omega_3\,\text{deg}(\Struct_i)+b_3),       & \text{S-MAR},  \\[-0.2em]
    \sigma(\omega_3\,\Struct_i^{(\text{MIS})}(h)+b_3), & \text{S-MNAR},
\end{cases} \\[0.3em]
&\text{N-MAR/MNAR: }\;
g(\Neigh_i(h))=
\begin{cases}
    \sigma(\omega_4\,\Neigh_i^{(\text{OBS})}(h)+b_4), & \text{N-MAR},  \\[-0.2em]
    \sigma(\omega_5\,\Neigh_i^{(\text{MIS})}(h)+b_5), & \text{N-MNAR}.
\end{cases}
\end{aligned}  
\end{equation}

Where, $\sigma(\cdot)$ is the sigmoid function, and $\text{deg}$ stands for \textit{degree}. The parameters $\omega_k$ and $b_k$ control the dependency strength and are calibrated to achieve the target probability $p_{\text{miss}}$. A detailed description of this calibration is provided in Section A.2 of the Supplementary Material.

\paragraph{Remark on S-MNAR.} While GAMM defines formally S-MNAR (see Eq.~\ref{eq:operational}), we exclude it from experiments as we assume a fully observed topology. Treating edges as missing shifts the problem to \textit{link prediction}. Furthermore, without edge masks to distinguish unobserved links from non-edges, the structural set $\Struct_i^{(\text{MIS})}(h)$ remains empty in our setting.

\paragraph{Imputation Methods.}
We evaluate a suite of state-of-the-art methods, focusing on those that represent current best practices. These methods have been selected because they demonstrate superior performance compared to older or less advanced baselines, thus we do not report standard baselines such as \textsc{MICE} \cite{van2011mice}.\\

\noindent\textbf{Tabular Imputers} process node attributes as a standard feature matrix, inherently ignoring graph structure: \textsc{Tabular Average} which imputes missing values with the global mean of each feature; \textsc{OT-tab} \cite{muzellecMissingDataImputation2020} that leverages optimal transport theory to minimize distributional discrepancies between observed and imputed data.

\noindent\textbf{Graph Imputers} explicitly exploit network topology for enhanced reconstruction. Notably we assess: \textsc{Graph Average}, a direct neighborhood average; \textsc{FP} \cite{rossiUnreasonableEffectivenessFeature2022} that propagates known values as a heat diffusion process along edges; \textsc{PCFI} \cite{um2023confidence} which introduces a novel channel-wise confidence metric for node features and utilizes shortest path distances for feature propagation and refinement; and finally \textsc{GRIOT} \cite{serrano2024reconstructing} which combines optimal transport with graph neural networks, to incorporates both node features and graph structure for robust imputation of missing attribute values.

\paragraph{Evaluation.}
We assess imputation quality using Mean Absolute Error (MAE) and Root Mean Square Error (RMSE). Since our primary contribution is a new missingness-generation taxonomy, these reconstruction metrics provide the most direct and controlled evaluation of how methods respond to the proposed mechanisms, avoiding confounding effects from downstream tasks. The relationship between reconstruction fidelity and downstream performance is known to be complex: downstream models may tolerate substantial missingness~\cite{serrano2024reconstructing}, and strong downstream accuracy does not necessarily imply reliable interpretability or high-quality imputation~\cite{shadbahr2022classification,shadbahr2023impact}. By focusing on reconstruction, we obtain a rigorous and targeted validation of our core contribution. All experiments are repeated eight times, and we use a Wilcoxon–Mann–Whitney U test~\cite{mann1947test}, with a significance level $p<0.05$ to assess statistical significance, testing the null hypothesis $H_0$ that the performance distributions of the compared methods are identical.

\subsection{Results and Analysis}
We evaluate state-of-the-art imputers under our taxonomy of missingness mechanisms, with a focus on the most challenging MNAR variants (Table~\ref{tab:diff_MNARquantileboth_vs_NMNAR_mae}), and summarise all 2,304 configurations in Table~\ref{tab:homophily_summary_stats_percentage}. 

\begin{table*}[t!]
\centering
\caption{Percentage degradation in MAE of N-MNAR over A-MNAR ($p_{\text{miss}}=20\%$) across dataset homophily groups. Negative results indicate deterioration in MAE with the N-MNAR mechanism compared to A-MNAR. (*) indicates statistical significance (Mann--Whitney U, $p < 0.05$).}
\label{tab:diff_MNARquantileboth_vs_NMNAR_mae}
\begin{center}
\begin{minipage}{0.4\linewidth}
\begin{center}
a) HOMOPHILIC datasets
\end{center}
\resizebox{\linewidth}{!}{
\begin{tabular}{lrrr}
\toprule
Imputer      & CiteSeer &     Cora &   PubMed \\ \midrule
Tabular\_Avg & {-50.6}* & {-48.7}* &  {-9.1}* \\
OT-tab       & {-10.3}* & {-14.0}* &  {-2.1}* \\ \midrule
FP           & {-72.9}* & {-68.9}* & {-15.5}* \\
GRIOT        &   -12.5~ & {-17.1}* &  {-5.1}* \\
Graph\_avg   & {-70.1}* & {-64.4}* & {-12.8}* \\
PCFI         & {-55.6}* & {-55.5}* & {-13.0}* \\ \bottomrule
\end{tabular}
}
\end{minipage}
\hfill
\begin{minipage}{0.5\linewidth}
\begin{center}
b) HETEROPHILIC datasets
\end{center}
\resizebox{\linewidth}{!}{
\begin{tabular}{lrrrr}
\toprule
Imputer      & Tolokers &  Cornell &    Texas & Wisconsin \\ \midrule
Tabular\_Avg &  {-9.0}* & {-47.5}* & {-46.5}* &  {-47.2}* \\
OT-tab       &  {-7.6}* & {-30.8}* & {-25.5}* &  {-31.2}* \\ \midrule
FP           & {-35.3}* & {-46.1}* & {-67.7}* &    -54.4~ \\
GRIOT        &    +0.9~ & {-77.1}* & {-72.3}* &  {-72.4}* \\
Graph\_avg   &  {-5.0}* & {-53.1}* & {-62.5}* &  {-54.3}* \\
PCFI         &  {-4.0}* & {-49.4}* & {-38.3}* &  {-48.9}* \\ \bottomrule
\end{tabular}
}
\end{minipage}
\end{center}

\begin{center}
\begin{minipage}{0.75\linewidth}
\begin{center}
c) NEUTRAL datasets
\end{center}
\resizebox{\linewidth}{!}{
\begin{tabular}{lrrrrr}
\toprule
Imputer      &    Actor & Chameleon & Squirrel & Minesweeper & Roman\_Empire \\ \midrule
Tabular\_Avg & {-53.1}* &  {-38.8}* & {-34.5}* &    {-23.1}* &        +11.8* \\
OT-tab       &  {-6.4}* &   {-6.2}* &  {-7.8}* &    {-20.1}* &         +3.1* \\ \midrule
FP           & {-73.3}* &  {-86.8}* & {-78.6}* &    {-27.4}* &         -0.6~ \\
GRIOT        & {-14.3}* &    -11.2~ &   -14.1~ &     {-0.0}* &         +3.3* \\
Graph\_avg   & {-63.5}* &  {-38.8}* & {-36.0}* &    {-26.4}* &         +6.2* \\
PCFI         & {-51.6}* &  {-17.2}* & {-17.4}* &    {-25.5}* &         +6.5* \\ \bottomrule
\end{tabular}
}
\end{minipage}
\end{center}
\end{table*}

\subsubsection{RQ1: Impact of Graph-Aware Mechanisms}
Across configurations, \textbf{graph-aware missingness markedly increases difficulty}. As captured in Table~\ref{tab:homophily_summary_stats_percentage}, moving from attribute-only to structurally informed masks leads to significant degradation in 46.6\% of cases, consistently across tabular and graph-based imputers. These results indicate that conventional attribute-only masks provide an overly optimistic view of robustness and that integrating structure in mask generation exposes realistic failure modes.

\begin{table*}[ht!]
\centering
\caption{Percentage Summary of Statistical Outcomes by mechanism. Total: 100\% = 2,304 experiments.}
\label{tab:homophily_summary_stats_percentage}
\resizebox{\linewidth}{!}{
\begin{tabular}{lclcccc}
\toprule
\multicolumn{3}{c}{Type}      & Sign. Degradation~~ & ~~No Sign. Change~~ & ~~Sign. Improvement~~ &  \textbf{Total}  \\ \midrule
A-MAR  & v.s. & N-MAR              &        8.0\%        &       32.2\%        &         2.9\%         & \textbf{43.1\%}  \\
A-MNAR & v.s. & N-MNAR             &       38.6\%        &        9.7\%        &         8.6\%         & \textbf{56.9\%}  \\ \midrule
\multicolumn{3}{c}{\textbf{Total}} &   \textbf{46.6\%}   &   \textbf{41.9\%}   &    \textbf{11.5\%}    & \textbf{100.0\%} \\ \bottomrule
\end{tabular}
}
\end{table*}

\subsubsection{RQ2: Role of Homophily under MAR}
Under MAR, the contrast between attribute-based (A-MAR) and neighborhood-based (N-MAR) mechanisms depends strongly on network homophily. In homophilic or neutral graphs, both mechanisms yield similar performance, as neighbors are either similar or redundant (Supplementary Material Section C). In heterophilic graphs, however, switching from A-MAR to N-MAR leads to a significant degradation for most approaches. This degradation arises because aggregating dissimilar neighbors injects misleading signals, highlighting the fragility of methods that implicitly rely on homophily.

\subsubsection{RQ3: MNAR with Neighborhood Dependence}    
The MNAR setting amplifies these effects. Missingness that depends on unobserved values interacts with topology in ways that make reconstruction substantially harder.
\\

\noindent\textit{Homophilic and Neutral Graphs.}
In these graphs, N-MNAR consistently outperforms A-MNAR in difficulty: Table~\ref{tab:diff_MNARquantileboth_vs_NMNAR_mae}(a,c) shows widespread and significant degradation when moving from A-MNAR to N-MNAR. For example, on Cora, \textsc{GRIOT} and \textsc{FP} incur a deterioration of 17.1\% and 68.9\%, respectively, for the MAE (both significant). Although the homophilic structure typically helps imputation, N-MNAR introduces dependencies between a node's missingness and its neighbors' unobserved values, making the reconstruction problem inherently more entangled than in A-MNAR.\\

\noindent\textit{Heterophilic Graphs.}
The effect is strongest in heterophilic graphs (Table~\ref{tab:diff_MNARquantileboth_vs_NMNAR_mae}(b)). For instance, \textsc{FP} presents a decrease of 35.3\% in MAE on Tolokers, and \textsc{GRIOT}'s performance decrease by 77.1\% on Cornell (p < 0.05). Here, dependence on unobserved values compounds with misleading neighbor information, creating a worst-case regime where all core assumptions of existing imputers are violated. This heterophilic N-MNAR combination emerges as a universal Achilles’ heel across methods and datasets.

\subsubsection{RQ4: Distributional Fidelity}
\begin{figure}[t]
\centering
\includegraphics[width=\linewidth]{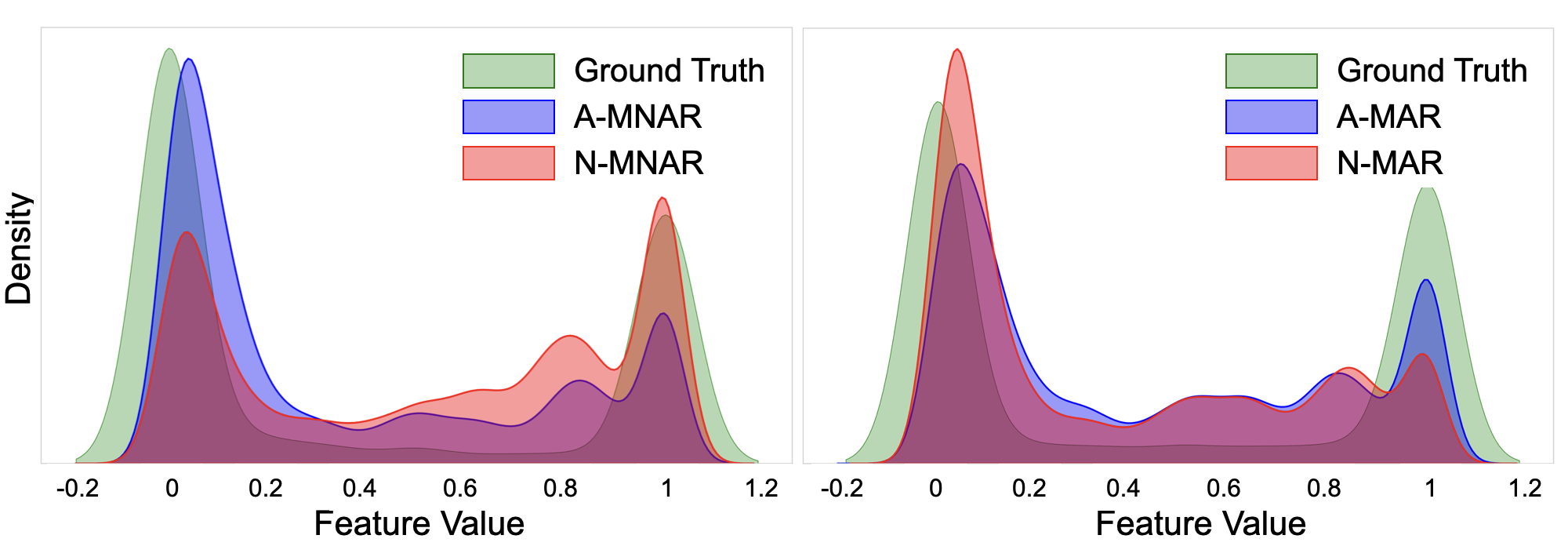}
\caption{Density (KDE) plots of imputed features on Tolokers with 20\% of missing values. MNAR (left); MAR (right)}
\label{fig:distribution}
\end{figure}

Pointwise errors do not guarantee correct distributional reconstruction. Figure~\ref{fig:distribution} illustrates distortions in binary attribute distributions on Tolokers (heterophilic) with 20\% missing values: imputers underestimate the mass at 0 in N-MNAR, and at 1 in A-MNAR, flattening the bimodal structure. This skew is visible under A-MAR but intensifies under N-MAR and becomes severe under N-MNAR. In these conditions, imputers face biased supervision (missingness depends on unobserved values) and misleading structural cues (dissimilar neighbors), jointly causing systematic underestimation and flattening.

\paragraph{Summary of Findings.}
\begin{sloppypar}
    We synthesise the main insights across RQs:
    \begin{enumerate}
        \item Graph-aware missingness markedly degrades imputation performance in nearly half of all configurations.
        \item Heterophily drives the MAR gap, with significant drops from A-MAR to N-MAR.
        \item Under MNAR, N-MNAR is consistently the hardest regime, especially on heterophilic graphs.
        \item Graph-aware mechanisms break distributional fidelity, introducing biases even when MAE/RMSE appear acceptable.
    \end{enumerate}
\end{sloppypar}

\section{Conclusion and Future Directions}\label{sec:conclusion}

\begin{sloppypar}
We introduced \textbf{GAMM}, a taxonomy extending traditional missing-data definitions to account for graph structure. Our experiments show that graph-aware missingness, especially neighbor-aware MNAR in heterophilic networks, can severely degrade imputation performance, and current state-of-the-art methods, designed for classical settings, struggle under these conditions. Although some downstream tasks remain robust, many rely on accurate feature reconstruction. We highlight the need for imputation methods resilient to these challenges, and for future work on missing structural data. By providing a theoretical framework, empirical evidence, and open-source code$\mathbf{^{\ref{ft:code}}}$, we aim to support more rigorous evaluation and development of imputation approaches for graphs.
\end{sloppypar}

\section{Acknowledgment}
This work has been partly funded by a public grant from the French National Research Agency (ANR) under the “France 2030” investment plan, which has the reference EUR MANUTECH SLEIGHT - ANR-17-EURE-0026.

\bibliographystyle{splncs04}
\bibliography{mybibliography}

\end{document}